\documentclass{article} 
\usepackage{iclr2025_conference,times}
\usepackage{tikz} 
\usepackage{listings}

\usepackage{wrapfig} 

\usepackage{pgfplots}
\usepackage{booktabs}
\usepackage{subcaption}
\usepackage{algorithm}
\usepackage{algpseudocode}
\usepackage{multirow}
\usepackage{makecell}   

\usepackage{amsmath,amsfonts,bm}

\def\eqref#1{equation~\ref{#1}}

\def\1{\bm{1}}

\DeclareMathAlphabet{\mathsfit}{\encodingdefault}{\sfdefault}{m}{sl}
\SetMathAlphabet{\mathsfit}{bold}{\encodingdefault}{\sfdefault}{bx}{n}

\usepackage{hyperref}
\usepackage{url}
\usepackage[most]{tcolorbox}
\usepackage{tcolorbox}

\usepackage{xcolor}
\usepackage{pifont}

\author{
Caia Costello$^{1, 2}$ \quad 
Simon Guo$^{1}$ \quad
Anna Goldie$^{1}$ \quad 
Azalia Mirhoseini$^{1}$ \\
$^{1}$Department of Computer Science, Stanford University \\
$^{2}$Ceramic AI \\
\texttt{caia@.stanford.edu}
}

\begin{document}

\title{Think, Prune, Train, Improve: \\
Scaling Reasoning Without Scaling Models}

\maketitle

\begin{abstract}
    Large language models (LLMs) have demonstrated strong capabilities in programming and mathematical reasoning tasks, but are constrained by limited high-quality training data. Synthetic data can be leveraged to enhance fine-tuning outcomes, but several factors influence this process, including model size, synthetic data volume, pruning strategy, and number of fine-tuning rounds. We explore these axes and investigate which conditions enable model self-improvement. We introduce the \textbf{Think, Prune, Train} process, a scalable framework that iteratively fine-tunes models on their own reasoning traces, using ground-truth pruning to ensure high-quality training data. This approach yields improved performance: on GSM8K, Gemma2-2B achieves a Pass@1 of 57.6\% (from 41.9\%), Gemma2-9B reaches 82\%, matching LLaMA-3.1-70B, and LLaMA-3.1-70B attains 91\%, even surpassing GPT-4o, demonstrating the effectiveness of self-generated reasoning and systematic data selection for improving LLM capabilities.
\end{abstract}

\section{Introduction}

State-of-the-art LLMs have been extensively trained on public text, yielding diminishing returns from additional web-scraped data. One promising approach is leveraging curated synthetic data to improve reasoning, an essential part of advancing code generation and mathematical problem-solving.

Recent frontier models like LLaMA 3.1 \cite{dubey2024llama3herdmodels} and DeepSeek R1 \cite{deepseek2024r1} demonstrate that post-training on reasoning traces coupled with supervised fine-tuning (SFT) on filtered (pruned) data works well to improve models. Their strong performance on coding and math benchmarks highlights how properly curated synthetic data can drive substantial performance gains. For smaller models such as LLaMA (1B, 3B) and Gemma (2B) (9B) \cite{gemmateam2024gemma2improvingopen}, distillation \cite{hinton2015distillingknowledgeneuralnetwork} coupled with fine-tuning on reasoning trace datasets has become the dominant post-training paradigm. However, distillation depends on the availability of larger models, while reasoning training relies on extensive external datasets.

This raises a fundamental question: \textbf{Can small models learn to reason using only self-generated data?} Prior attempts at recursive fine-tuning on unfiltered text data \cite{shumailov2024nature} have observed {\em model collapse}: model degradation, including knowledge forgetting \cite{doi:10.1073/pnas.1611835114} and hallucination. However, those works explored simple text generation, a task without a ground truth metric for pruning. In contrast, \textit{mode collapse} refers to a model converging on a narrower set of high-probability outputs, reducing diversity at the cost of exploration. 
Mode collapse and model collapse are distinct yet interconnected risks in iterative fine-tuning. Our approach, recursive SFT on self-generated reasoning traces, might seem prone to mode collapse, as models increasingly favor confident, correct solutions, but results show that while Pass@1 improves significantly, Pass@50 and Pass@20 remain relatively stable, indicating that the model maintains diversity while prioritizing reliability. By incorporating correctness-based pruning, we aim to avoid both model and mode collapse while leveraging the benefits of iterative refinement for scalable reasoning improvement.

We investigate the conditions that enable self-improvement, particularly in smaller models, to understand when and how models can refine their reasoning abilities without external supervision. Our analysis demonstrates that reasoning trace selection through pruning can yield significant performance gains. Several factors influence this process, including the \textbf{size of the model} used for generating data, the \textbf{amount of synthetic data} incorporated during training, the \textbf{pruning strategy} employed for selecting reasoning traces, and the \textbf{number of fine-tuning rounds} conducted throughout the improvement process. We categorize this process into three components:  \textbf{Think, Prune, Train}, where models are iteratively fine-tuned on their own \textit{correct} step-by-step reasoning solutions.

\begin{enumerate}
\item Prompt models to reason in a structured way.
\item Prune incorrect outputs using ground-truth correctness filtering. 
\item Perform SFT on the current model with its own unique validated solutions.
\end{enumerate}





\begin{figure}[ht]
    \centering
    \includegraphics[width=0.7\textwidth]{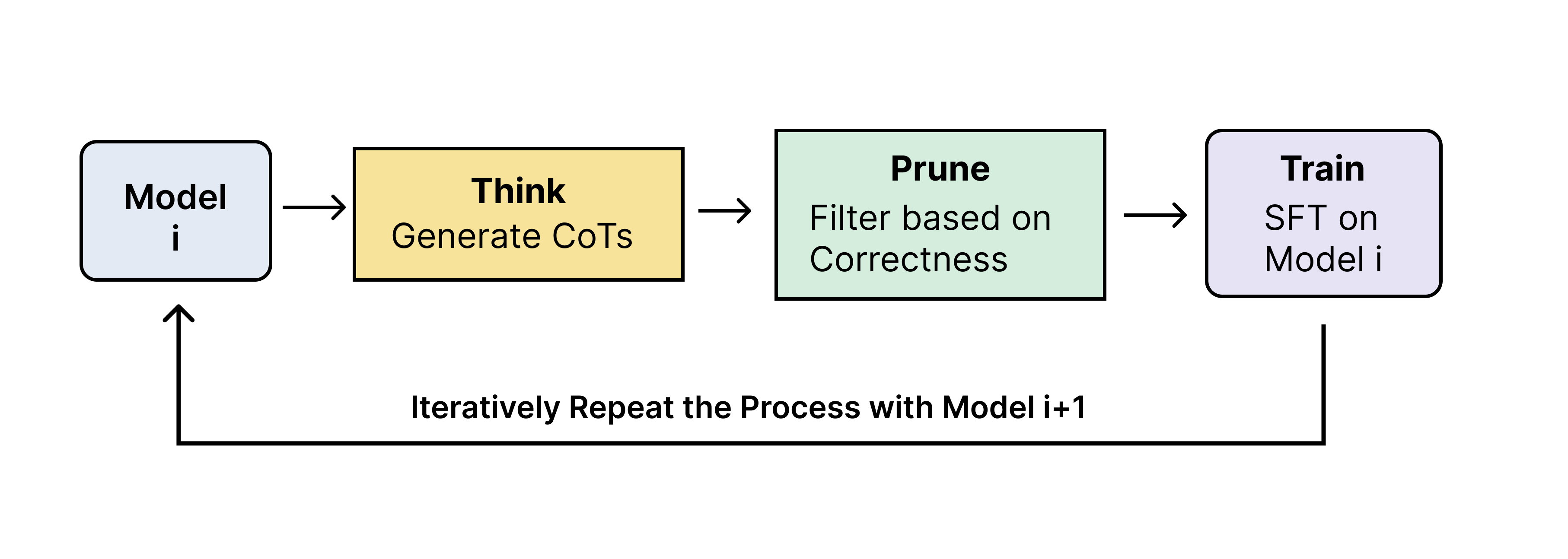} 
    \caption{\textbf{Recurrent process for model training.} Data generation, pruning, and supervised fine-tuning (SFT), with the arrow indicating the feedback loop to the tuned model.}
    \label{fig:process_flow_recurrent}
\end{figure}

By applying our process, as illustrated in Figure \ref{fig:process_flow_recurrent} , on small models, they achieve reasoning performance comparable to larger ones on coding and math benchmarks. In particular, on GSM8K \cite{cobbe2021trainingverifierssolvemath}, and CodeContests \cite{Li_2022cc}, we achieve the following performance gains:
\begin{itemize}
\item On GSM8K, Gemma2-2B improves from 41.9\% to 57.6\% for Pass@1
\item On GSM8K, Gemma2-9B reaches 82\% Pass@1, surpassing LLaMA3.1-70B-Instruct’s 78\% for Pass@1.
\item On GSM8K, LLaMA3.1-70B climbs from 78\% to 91\% Pass@1, outperforming even GPT-4o (2024-08-06).
\item On Code Contest, Pass@1 improves from 0.90\% to 1.14\% for the Gemma2-2B model and from 5.10\% to  7.90\% for the Gemma2-9B model.
\end{itemize}

\textbf{Our Contributions:}
\begin{enumerate}
\item \textbf{Avoids model collapse in self-training}: Prior works show that unfiltered recursive fine-tuning can degrade model performance, leading to catastrophic forgetting and hallucination. We demonstrate that \textbf{correctness-based pruning} stabilizes training, preserving knowledge while enhancing reasoning.
\item \textbf{ Insights for iterative reasoning refinement}: We analyze how data volume, model size, pruning strategies, and fine-tuning iterations influence self-improvement and identify the conditions for effective self-improvement. 
\item \textbf{Think, Prune Train framework}: We evaluate the Think, Prune, Train\textbf{ (TPT)} framework, demonstrating that structured reasoning prompting, correctness-based pruning, and supervised fine-tuning on validated solutions enables self-improvement.
\end{enumerate}

\section{Related work}
Prior work on improving reasoning in small models includes distillation using synthetic data from larger models, self-improvement strategies, and reinforcement learning (RL)-based optimization. In contrast, we evaluate whether self-improvement can be achieved without relying on teacher models, complex frameworks, or RL.

\textbf{Distillation:}

Previous work has shown success in fine-tuning base models with synthetic data from large models, a form of distillation: Alpaca \cite{taori2023alpaca}, MagicCoder \cite{wei2024magicoderempoweringcodegeneration}, and CodeGemma \cite{codegemmateam2024codegemmaopencodemodels} demonstrate effective single-round supervised fine-tuning (SFT) using data from larger models. 
Research in math reasoning has focused on structured solutions and synthetic data:
GSM8K \cite{cobbe2021gsm8k} provides high-quality grade-school word problems requiring step-by-step solutions, though we found that training solely on this dataset proved insufficient for scaling performance.
TinyGSM \cite{liu2024tinygsm} creates 12.3M synthetic math problems using GPT-3.5-turbo, achieving 63.1\% pass@1 even with a 125M parameter model. 
MetaMath \cite{lewkowycz2022solvingquantitativereasoningproblems} demonstrates the effectiveness of training models on multi-step forward and backward reasoning traces generated by GPT-3.5-Turbo, highlighting the effect of structured problem-solving data in fine-tuning outcomes. 

\textbf{Self-Improvement:}

Achieving iterative self-improvement using data from the same model presents a challenge. LLaMA 3.1 \cite{dubey2024llama3herdmodels} implements multi-round fine-tuning and reports that iterative training on synthetic data from their 405B model improves its 8B and 70B counterparts, but training on its own outputs without pruning led to performance degradation. To address this, they adopted a  pipeline consisting of supervised fine-tuning (SFT), followed by direct preference optimization (DPO), additional data generation, and data pruning. ReST \cite{uesato2023reinforced} proposes an offline RL framework that optimizes model behavior using a growing batch of self-generated training data pruned by a reward model, their algorithm aligns the language model’s outputs with human preferences, which are modeled using a learned reward function. Self-Instruct, \cite{wang2023selfinstructaligninglanguagemodels} introduces a scalable method for bootstrapping instruction-following models without requiring human-annotated datasets, by prompting GPT3 to generate diverse instruction-response pairs. Self-Taught Reasoner (STaR) \cite{zelikman2022star} proposes an iterative fine-tuning approach where a model learns to generate its own rationales while progressively improving its problem-solving ability.  STaR constructs rationalized solutions through rejection sampling and fine-tunes a base model on both correct and corrected (giving the model answer working backwards) rationales. 

\textbf{Reinforcement Learning:}

Reinforcement learning (RL) has played a significant role in improving the reasoning, decision-making, and alignment of large-scale language models. Traditional RL-based approaches such as Proximal Policy Optimization (PPO) \cite{schulman2017proximal} and Reinforcement Learning from Human Feedback (RLHF) \cite{ouyang2022instructgpt} have been widely used to fine-tune models for better instruction-following and alignment. 
However, newer RL methods have introduced refinements. 
Vine-PPO \cite{wang2024vine} enhances policy optimization by incorporating variance reduction techniques, improving convergence stability and sample efficiency. DeepSeek R1 \cite{deepseek2024r1} applies RL-based rejection sampling to filter and refine reasoning paths. Their pipeline first pre-trains a model with structured reasoning objectives, followed by RL optimization to improve output quality. ReST-EM \cite{uesato2024restem} extends ReST by incorporating Expectation Maximization, dynamically adjusting the weight of self-generated training data based on model confidence. Unlike RL-based methods like DeepSeek R1 and LLaMA 3.1 \cite{dubey2024llama3herdmodels} we focus exclusively on recursive correctness-based fine-tuning, demonstrating that self-improvement is possible without RL by leveraging structured prompting and validation.

\begin{table}[h]
\centering
\scriptsize
\renewcommand{\arraystretch}{1.0} 
\setlength{\tabcolsep}{4pt} 

\begin{tabular}{lcccc}
\toprule
\textbf{Approach} & \textbf{Uses CoT} & \makecell{\textbf{Only Ground Truth} \\ \textbf{Pruning}} & \makecell{\textbf{Starts from} \\ \textbf{Fine-tune}} & \makecell{\textbf{Multiple} \\ \textbf{Iterations}} \\
\midrule
\textbf{TPT (Ours)} & \checkmark & \checkmark & \checkmark & \checkmark \\
\midrule
\textbf{LLaMA 3.1} \cite{dubey2024llama3herdmodels} & \checkmark & \ding{55} & \checkmark & \checkmark \\
\textbf{DeepSeek R1} \cite{deepseek2024r1} & \checkmark & \ding{55} & \ding{55} & \checkmark \\
\textbf{ReSTem} \cite{huang2023beyond} & \ding{55} & \checkmark & \ding{55} & \checkmark \\
\textbf{MagicCoder} \cite{wei2024magicoderempoweringcodegeneration} & \ding{55} & \ding{55} & \ding{55} & \ding{55} \\
\textbf{Self-Instruct} \cite{wang2023selfinstructaligninglanguagemodels} & \ding{55} & \ding{55} & \ding{55} & \ding{55} \\
\textbf{STaR} \cite{zelikman2022star} & \checkmark & \checkmark & \ding{55} & \checkmark \\
\bottomrule
\end{tabular}

\caption{\textbf{Comparison of TPT with Related Approaches}. Compares if each approach uses Chain-of-Thought (CoT), applies ground truth pruning, starts from a fine-tuned model,  as opposed to the base model, and performs multiple fine-tuning iterations.}
\label{tab:tpt_comparison}
\end{table}

\subsection{SFT on Pruned Data as a Special Case of Policy Gradient}

Recent work suggests that supervised fine-tuning (SFT) and reinforcement learning (RL) share key similarities. \textsc{ReST}\textsubscript{EM} \cite{huang2023beyond} reframes RL as an Expectation-Maximization (EM) process, where policy updates occur on fixed data rather than dynamically evolving policies. This aligns with our approach, which fine-tunes on correctness-filtered data without explicit policy optimization. In policy gradient RL, which optimizes a parameterized policy $\pi_{\theta}$ to maximize expected return $J(\pi_{\theta})$:

\[\nabla J(\pi_{\theta}) = \mathbb{E}_{\tau \sim \pi_{\theta}} \Bigg[ \sum_{t=0}^{T} \nabla_{\theta} \log \pi_{\theta}(a_t | s_t) R(\tau) \Bigg]
\]

For language modeling, states correspond to token sequences $s_t = (x_0, ..., x_t)$ and actions to next-token choices $a_t = x_{t+1}$. Sparse rewards $R(\tau)$, such as binary correctness in code generation, are $-1$ for negative $x^l$ examples and $+1$ for positive ones $x^w$.

Because errors in math problems are localized we notice that the gradient of the negative example is of equal magnitude but opposite sign to the positive example on those steps where they differ.  Under this assumption, separating out the positive from the negative examples,  the gradient simplifies to:
$$
\nabla_{\theta} \log \pi(x^w_{t+1} | x_{0:t}) - \nabla_{\theta} \log \pi(x^l_{t+1} | x_{0:t}) $$
$$
= 2 \nabla_{\theta} \log \pi(x^w_{t+1} | x_{0:t}).
$$  
This mirrors the standard cross-entropy loss in SFT for just the positive examples:
\[
L(\pi_{\theta}) = - \sum_{t=0}^{T} \log \pi_{\theta}(x_{t+1} | x_{0:t})
\]

Thus, filtering for correctness and fine-tuning on self-generated sequences approximates policy gradient updates. This suggests that SFT on correct outputs is not just a heuristic but an implicit RL formulation, where correctness serves as a reward signal guiding model improvement.

\section{Think, Prune, Train:}

We systematically investigate the key components that enable effective self-improvement without dependence on external models or datasets. Many variables contribute to the effectiveness of self-improvement techniques, such as:

\begin{itemize}
    \item \textbf{Choice of Base Model:} We focus on the Gemma and Llama families, particularly smaller models.
    \item \textbf{Prompting Strategy:} Utilize CoT prompting with a temperature of 0.8 to generate data from the training sets.
    \item \textbf{Dataset Size:} Explore how varying amounts of synthetic data impact fine-tuning outcomes.
    \item \textbf{Comparison with Distillation:} We compare the performance of self-generated data with data generated by larger models.
    \item \textbf{Fine-Tuning Rounds:} Examine the effects of fine-tuning the base model once versus iteratively fine-tuning over multiple rounds
    \item \textbf{Pruning Strategy:} Compare the performance of selecting data based on ground truth, partially correct solutions, and using unfiltered generated solutions. 
\end{itemize}

While prior work often formulates self-improvement as a reinforcement learning (RL) problem, requiring explicit reward models and policy optimization. We apply supervised fine-tuning (SFT) to a model using its own generated outputs, selectively filtering for correctness. Our analysis explores whether iterative refinement can be achieved purely through structured data selection.

\subsection{Choosing Data for Self-Improvement}

A key challenge in evaluating iterative self-improvement is disentangling gains due to actual refinement of the model's reasoning capabilities from those caused by training on an increasingly large accumulated dataset \cite{gerstgrasser2024modelcollapseinevitablebreaking}. To systematically investigate this, we explored different data retention strategies across fine-tuning rounds. Initially, we considered accumulating multiple solutions per question over successive rounds, allowing the model to learn from a more diverse set of responses. While this yields performance improvements, the model's enhancement could be attributed to data size augmentation rather than genuine iterative refinement. 

To isolate the effects of self-improvement, we retain only a single, \textit{unique} example per question per round, ensuring that the dataset size remains constant across iterations. Moreover, instead of accumulating data across rounds, we replaced each round’s dataset entirely with newly generated solutions, so every model is strictly trained on \textit{only} self-generated data. Our findings suggest that with strict data constraints, iterative fine-tuning can lead to meaningful gains, demonstrating that model improvement is not solely a function of dataset expansion. Our experiments converged to a simple yet effective method for model self-improvement, illustrated in Algorithm \ref{recursive-algorithm}.

\subsection{How TPT is different}

Prior self-improvement methods, such as ReST \cite{uesato2023reinforced} and the post-training setup of LLaMA 3.1 \cite{dubey2024llama3herdmodels}, refine models iteratively by applying their process to an already fine-tuned model. ReST prunes self-generated data using a learned reward function, while LLaMA 3.1 applies direct preference optimization (DPO) and prunes data using both reward models and ground-truth filtering.

STaR \cite{zelikman2022star} uses rejection sampling and fine-tunes the base model on both correct and corrected rationales, incorporating backward reasoning where the model reasons from a given answer. However, STaR does not iteratively fine-tune on its improved model.

Our approach focuses solely on supervised fine-tuning (SFT) to examine whether a model can iteratively improve from its own generated outputs using correctness-based filtering as the only selection mechanism. By focusing soley on SFT, while forgoing, RL, reward models, and external correction, we isolate the role of structured data selection in self-improvement.

\begin{algorithm}
\caption{Think, Prune, Train}
\begin{algorithmic}
\State \textbf{Input:} Base model $M_0$, pruning strategy $P$, amount of data $|n| = f$ used for supervised fine-tuning
\For{$i = 1$ to $N$}
    \State SFT $M_{i-1}$ on pruned data $f_{i-1} \rightarrow M_i$
    \State Generate 10 solutions for all problems in the train set using model $M_i$ to construct $S$, the set of all generated solutions
    \State Apply pruning strategy $P$ to all new solutions $S$
    \State Randomly sample $f_i \subset S$ as examples for fine-tuning on $M_i$
\EndFor
\State \textbf{Output:} Improved model $M_N$
\end{algorithmic}
\label{recursive-algorithm}
\end{algorithm}






\subsection{Experimental Setup}

We study the effectiveness of instruction fine-tuned variants of the Llama and Gemma model families. Specifically, we experiment with \texttt{gemma2-2b-it}, \texttt{gemma2-9b-it}, \texttt{Llama-3.2-1B-Instruct}, and \texttt{Llama-3.3-70B-Instruct}. Our SFT runs use a learning rate of \texttt{1e-6} for Gemma and \texttt{1e-5} for Llama to maintain training stability. We employ the AdamW optimizer \cite{loshchilov2017decoupled} with default weight decay. Training is conducted over a single epoch with a 10\% warm-up of training steps to mitigate overfitting.

The appendix (\ref{appendix-traces}) contains synthetic example solutions from question 20 of the GSM8K \cite{cobbe2021gsm8k} train set, illustrating the reasoning steps taken by different models. For the iterative fine-tuning experiments using Algorithm~\ref{recursive-algorithm}, we set the number of pruned examples per round to 2000 for GSM8K and 1000 for CodeContests, performing up to $N=4$ rounds of iterative improvement. Evaluation is conducted at a temperature of 0.7. For GSM8K, we select a subset of 500 questions from the test set, ensuring that answers exactly match the ground truth, including formatting, to enforce both correctness and adherence to the expected answer format. Additionally, we evaluate our recursively trained models on the 140 questions without image tags from CodeContests, using standard input/output matching to verify correctness.

\section{Results: }

Our results are broken into three sections: 

\begin{itemize}
    \item We first analyze how dataset size and the model generating the data impact fine-tuning outcomes.
    \item Next, we evaluate the Think-Prune-Train process on GSM8K and CodeContests
    \item Finally, we validate our pruning strategy, demonstrating that pruned synthetic data outperforms unpruned data and that our approach generalizes to LLaMA variants.
\end{itemize}

\subsection{Synthetic data scaling }

\textbf{We investigate whether increasing the size of synthetic datasets, and the models used to generate them, leads to performance gains.}  
Experimental results reveal that more synthetic data does not necessarily translate to better outcomes across all dataset types. We also compare how SFT on increasing amounts of human generated data from the training set effects the model's performance for comparison.

For GSM8K, we observe that small amounts of synthetic data generated by the larger Gemma-9B model outperform those from the 2B model. At 2,000 examples, the 9B-generated data achieves a \textbf{Pass@1} of \textbf{54.0\%}, compared to \textbf{52.5\%} for 2B-generated data. However, as dataset size increases, performance plateaus or even slightly declines for both 2B and 9B-generated data.  

\textbf{\textit{The results indicate that simply scaling data is not a universally effective strategy.}} performance is shaped by the interplay between data quality, dataset size, and the model used for generation.

\begin{table}[h]
\centering
\tiny 
\setlength{\tabcolsep}{1pt} 
\renewcommand{\arraystretch}{0.7}\begin{tabular}{llcccc}
\toprule
\textbf{Model} & \textbf{Approach} & \textbf{Size} & \textbf{Data Source} & \textbf{Pass@1(\%)} & \textbf{Pass@20(\%)} \\
\midrule
\multirow{1}{*}{\textbf{Gemma-2-2B}}  
& Baseline & - & - & 41.9 & 76.0  \\
& Self-gen & 1k & 2B & 50.6 & 81.8  \\
&  & 2k & 2B & 52.5 & \textbf{83.0}  \\
&  & 4k & 2B & \textbf{54.8} & 82.6  \\
&  & 6k & 2B & 51.6 & 81.0  \\
& Human & 2k & GSM8K & 45.5 & 83.0  \\
&  & 4k & GSM8K & 44.3 & 83.4  \\
&  & 6k & GSM8K & 44.0 & 85.6  \\
& Distill & 1k & 9B & 51.2 & 83.1  \\
& & 2k & 9B & \textbf{54.0} & \textbf{83.8}  \\
&  & 4k & 9B & 53.1 & 83.0  \\
&  & 6k & 9B & 51.2 & 82.6  \\
\midrule
\multirow{2}{*}{\textbf{Gemma-2-9B}}  
& Baseline & - & - & 66.4 & 88.4  \\
& Self-gen & 1k & 9B & 80.5 & 89.0  \\
&  & 2k & 9B & 81.1 & 90.0  \\
\bottomrule
\end{tabular}
\caption{\textbf{Scaling synthetic dataset size shows no clear trends, motivating a different solution (TPT) \ref{recursive-algorithm}.}  
GSM8K Pass@1\% and Pass@20\% for fine-tuned  Gemma-2-2B and Gemma-2-9B models scaling number of training examples.}
\label{tab:gsm8k_math_performance}
\end{table}

\subsection{Recursive data performance }

\textbf{Next, we examine how recursive synthetic data generation influences performance across multiple tasks.} Our investigation reveals a pattern of cumulative improvements across multiple tasks, particularly for single-attempt performance. These findings align with \cite{muennighoff2025s1simpletesttimescaling}, though with a key distinction: while their approach begins with models explicitly trained to reason and uses budget forcing to extend reasoning length, our method starts with plain instruct models and prompts them to develop reasoning behaviors.

\textbf{Mathematical Reasoning Recursive Performance:}
Through recursive fine-tuning, where each subsequent model (Model1, Model2, etc.) is trained on synthetic data generated by its predecessor, we observe significant gains in mathematical reasoning capability. The Gemma-2B model's exact match Pass@1 performance on GSM8K, as seen in Table \ref{tab:gsm8k_math_performance} ,increases from 41.9\% to 57.6\% over four iterations of recursive training, accompanied by  gains in Correct@20. Similarly, the Gemma-9B model demonstrates improvement, reaching a Pass@1 of 82.4\% from 66.4 \% within three iterations. 

While Pass@1 steadily increases, \textbf{Pass@20 shows diminishing returns, mostly plateauing after the first iteration,} suggesting that recursive training primarily enhances accuracy rather than improving diversity across sampled generations.

We hypothesize that the reasoning-intensive nature of mathematical tasks enables recursive fine-tuning to build upon previous patterns.  This phenomenon aligns with findings from STAR \cite{zelikman2022star}, which highlights the effectiveness of structured rationale generation for reasoning tasks.

\begin{figure}[ht]
\centering
\begin{minipage}[c]{0.4\textwidth}
\small
\captionof{figure}{\textbf{Recursive training enhances GSM8K Pass@1/20 performance in Gemma models.}  On GSM8K Gemma2-2B model’s Pass@1(\%) performance increases from 41.9\% to 57.6\% over four iterations of this process \ref{recursive-algorithm} While, the Gemma-9B model improves to a Pass@1 of 82.4\%. Bars M1-4 represent the models trained though the TPT process, starting with base Gemma2-2B/9B . }
\label{fig:performance_comparison_gsm8krecur}
\end{minipage}%
\hfill
\begin{minipage}[c]{0.55\textwidth}
\centering
\includegraphics[width=\linewidth]{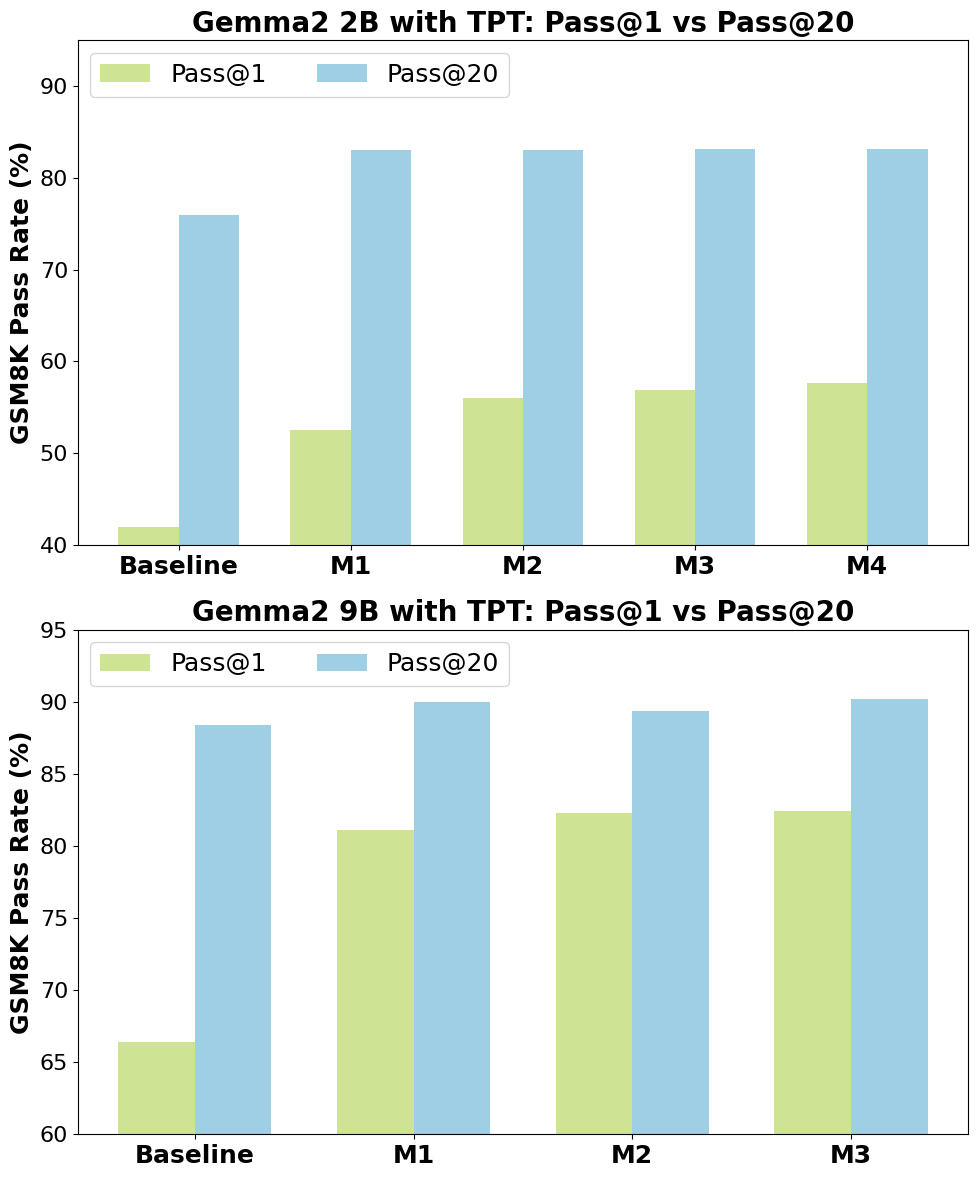}
\end{minipage}
\end{figure}

\begin{table*}[t]
\tiny
\setlength{\tabcolsep}{1.2pt}
\renewcommand{\arraystretch}{0.8}

\begin{minipage}[t]{0.48\textwidth}
\centering
\begin{tabular}{llccccc}
\toprule
\textbf{Model} & \textbf{Stage} & \textbf{Source} & \textbf{Data} & \textbf{Pass@1} & \textbf{Pass@20} & \textbf{Correct@20} \\
\midrule
\multirow{1}{*}{\textbf{Gemma-2-2B}} 
& Baseline & - & - & 41.9 & 76.0 & 380 \\
& Init (Model 1) & 2B & 2k & 52.5 & 83.0 & 415 \\
& Rec 1 (Model 2) & M1 & 2k & 56.0 & 83.0 & 415 \\
& Rec 2 (Model 3) & M2 & 2k & 56.9 & 83.2 & 416 \\
& Rec 3 (Model 4) & M3 & 2k & \textbf{57.6} & \textbf{83.2} & 416 \\
\midrule
\multirow{1}{*}{\textbf{Gemma-2-9B}} 
& Baseline & - & - & 66.4 & 88.4 & 442 \\
& Init (Model 1) & 9B & 2k & 81.1 & 90.0 & 450 \\
& Rec 1 (Model 2) & M1 & 2k & 82.3 & 89.4 & 447 \\
& Rec 2 (Model 3) & M2 & 2k & \textbf{82.4} & \textbf{90.2} & 451 \\
\bottomrule
\end{tabular}
\caption{\textbf{Recursive fine-tuning improves GSM8K performance across multiple iterations.} 
Performance of recursively trained Gemma models on GSM8K, showing Pass@1(\%) and Pass@20(\%) improvement when following the algorithm, These are the numbers for Figure \ref{fig:performance_comparison_gsm8krecur}  Pass@1 shows a clearer upward trend with Pass@20 plateauing after 2 rounds. }
\label{tab:math_recursive_performance}
\end{minipage}%
\hspace{0.02\textwidth}
\begin{minipage}[t]{0.48\textwidth}
\centering
\begin{tabular}{llccccc}
\toprule
\textbf{Model} & \textbf{Stage} & \textbf{Source} & \textbf{Data} & \textbf{Pass@1} & \textbf{Pass@50} & \textbf{Correct@50} \\
\midrule
\multirow{1}{*}{\textbf{Gemma-2-2B}} \\ (CodeContest) 
& Baseline & - & - & 0.90 & 5.70 & 8 \\
& Init (Model 1) & 2B & 1k & 0.84 & 7.14 & 10 \\
& Rec (Model 2) & M1 & 1k & 1.00 & 9.20 & 13 \\
& Rec (Model 3) & M2 & 1k & 1.10 & \textbf{9.20} & 13 \\
& Rec (Model 4) & M3 & 1k & \textbf{1.14} & 7.85 & 11 \\
\midrule
\multirow{1}{*}{\textbf{Gemma-2-9B}} \\ (CodeContest)
& Baseline & - & - & 5.10 & 15.00 & 21 \\
& Init (Model 1) & 9B & 1k & 7.30 & 18.50 & 26 \\
& Rec (Model 2) & M1 & 1k & 7.60 & 17.80 & 25 \\
& Rec (Model 3) & M2 & 1k & \textbf{7.90} & \textbf{18.50} & 26 \\
\bottomrule
\end{tabular}
\caption{\textbf{Recursive Fine-Tuning on Code Contest dataset shows minor improvement at Pass@1 and plateaus for Pass@50.}  
Pass@1 improves steadily through iterative fine-tuning, reaching 1.14\% for the 2B model and 7.90\% for the 9B model. However, Pass@50(\%) exhibits diminishing returns, indicating potential limitations in further refinement. The Correct@50 metric further illustrates that while improvements occur, gains diminish over iterations.}
\label{tab:code_contest_synthetic}
\end{minipage}
\end{table*}

\textbf{Code Contest Recursive Results: }
In the domain of code generation, we observe more nuanced outcomes, \textbf{Code Contest pass@1 performance improves but pass@50 plateaus}.
The 2B model steadily improves in Code Contest, seen in Table \ref{tab:code_contest_synthetic} Pass@1, rising from 0.9\% to 1.14\% after four recursive training iterations. However, Pass@50 plateaus around 7.85–9.2\%, possibly due to code generation being less reliant on step-by-step reasoning. We also trained on 1k data each recursive round as opposed to 2k for GSM8K, this is due to the smaller dataset size, as well as lower model accuracy  on that dataset.

\subsection{Impact of Pruning on Synthetic Data Quality}
\textbf{Pruning plays a crucial role in ensuring the effectiveness of self-generated synthetic data} directly influencing model performance. Experiments with un-pruned self-generated data showed deteriorated performance, contrasting with \cite{setlur2024rlincorrectsyntheticdata} which suggests value in incorrect data for mathematical tasks when critical errors are avoided. With pruning, models trained on 2B and 9B-generated data achieve comparable performance. However, without pruning, 9B-generated data significantly outperforms 2B-generated data across all tasks, demonstrating pruning's role in equalizing data quality across model sizes. We also show results from soft pruning experiments where we keep examples that pass one or more tests where we observed increased diversity. 

\begin{table}[h]
\centering
\scriptsize 
\setlength{\tabcolsep}{1pt} 
\renewcommand{\arraystretch}{0.80} 
\begin{tabular}{llcccc}
\toprule
\textbf{Model} & \textbf{ Stage} & \textbf{ Source} & \textbf{ Data} & \textbf{Pass@1(\%)} & \textbf{Pass@20/50(\%)} \\
\midrule
\multirow{1}{*}{\textbf{Gemma-2-2B}} \\ (GSM8K)
& Baseline & - & - & 41.9 & 76.0  \\
& 2k No-Prune & 2B & 2k & 45.13 & 71.6  \\
& 6k No-Prune & 2B & 6k & 44.57 & 71.0  \\
\midrule
\multirow{1}{*}{\textbf{Gemma-2-2B}} \\ (Code Contest)
& Baseline & - & - & 0.90 & 5.70  \\
& 1k No-Prune & 2B & 1k & 0.85 & 5.70  \\
& 1k No-Prune (9B Data) & 9B & 1k & 0.94 & 6.40  \\
& 5k No-Prune & 2B & 5k & 0.90 & 6.42  \\
& 5k No-Prune (9B Data) & 9B & 5k & 0.67 & 9.29  \\
& 1k Soft-Pos & 2B & 1k & 0.70 & 8.57  \\
\bottomrule
\end{tabular}
\caption{\textbf{SFT without pruning performs poorly, while soft pruning increases diversity.}  
Pass@1 and Pass@20 and Pass@50, results for GSM8K and Code Contest respectively on Gemma-2-2B models trained on different dataset sizes and pruning strategies.}
\label{tab:noprune}
\end{table}

\subsection{Results synthesis:}

Our experimental results demonstrate differences in learning efficiency across model scales when applying TPT (Think Prune Train) recursive fine-tuning. The smaller Gemma-2-2B model required four complete recursive TPT rounds to achieve its peak accuracy, while the medium-sized Gemma-2-9B improved faster after three rounds. In contrast, the significantly larger LLaMA-70B model attained 91.5\% pass@1 accuracy after merely a single round of TPT recursive fine-tuning. The enhanced Pass@1 metric is noteworthy as it measures the model's ability to generate correct solutions on the first attempt. LLaMA-70B's 91.5\% pass@1 score indicates that TPT effectively transfers reasoning abilities even with minimal self-generated training iterations when applied to larger model architectures.These findings highlight the importance of considering model parameter size when designing training regimes.

\begin{figure}[h]
\centering
\begin{minipage}[c]{0.43\textwidth}
\small
\captionof{figure}{\textbf{Recursive Fine-Tuning Can Improve Over Open Source Models.}  
The \texttt{Gemma-2-9B} model improves from 66.0\% to 82.4\%, while \texttt{LLaMA-70B} improves from 76.0\% to 91.5\% Pass@1, surpassing \texttt{GPT-4o} (82\%). This figure highlights the performance gains achieved through recursive fine-tuning.}
\label{fig:pass1_gsm8k}
\end{minipage}%
\hfill
\begin{minipage}[c]{0.55\textwidth}
\centering
\includegraphics[width=\linewidth]{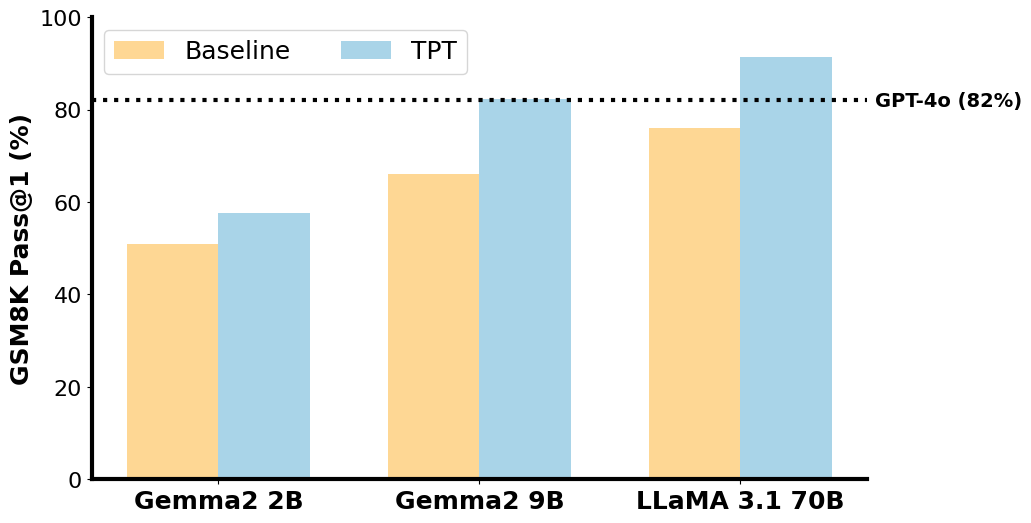}
\end{minipage}
\end{figure}

\begin{table}[h]
\centering
\scriptsize 
\setlength{\tabcolsep}{2pt} 
\renewcommand{\arraystretch}{0.85} 
\begin{tabular}{llcccc}
\toprule
\textbf{Model} & \textbf{ Stage} & \textbf{Source} & \textbf{Data} & \textbf{Pass@1(\%)} & \textbf{Pass@20(\%)} \\
\midrule
\multirow{1}{*}{\textbf{LLaMA-1B (GSM8K)}}  
& Baseline & - & - & 18.2 & 65.2  \\
& 2k Fine-Tune & LLaMA-1B & 2k & 20.8 & 68.4  \\
\midrule
\multirow{1}{*}{\textbf{LLaMA-70B (GSM8K)}}  
& Baseline & - & - & 78.6 & 92.2  \\
& 6k Fine-Tune & LLaMA-70B & 6k & \textbf{91.5} & \textbf{95.8}  \\
\bottomrule
\end{tabular}
\caption{\textbf{The Think-Prune-Train method improves GSM8K performance on LLaMA models}  
One round of the TPT process improves performance. Notably, LLaMA-70B achieves a Pass@1 increase from 78.6\% to 91.5\%}
\label{tab:llama_gsm8k}
\end{table}

\subsection{ Possible mode collapse  }
\label{modecollapse}
Our observed Pass@1 improvements alongside stable Pass@20 and Pass@50 metrics suggest a form of \textit{mode collapse} \cite{9934291} in recursive synthetic data generation, where the model prioritizes high-confidence solutions over diversity. Mode collapse occurs when a model converges on a narrower set of high-probability outputs at the expense of diversity. This phenomenon has been studied in recent work on synthetic data feedback loops, including \cite{gerstgrasser2024modelcollapse}, which explores the long-term effects of training models on their own synthetic outputs. 

The authors argue that recursive training pipelines without sufficient real and synthetic data accumulation can lead to progressively narrowing distributions, where models produce increasingly homogeneous outputs, and performance degradation. While their work emphasizes the dangers of such feedback loops—particularly in open-ended, creative tasks—our setting differs in that correctness is strictly defined. We do observe less diversity in outputs, but increased correctness, this outcome is not inherently detrimental, particularly for domains such as mathematics and programming, where correctness and efficiency take precedence over diversity.

Though we observe reduced output variety, increased accuracy aligns with our goal of enhancing precision through pruning. This effect may relate to fine-tuning’s impact on hallucinations \cite{kang2024unfamiliarfinetuningexamplescontrol}. If diversity were the priority, techniques like soft-positive sampling, as seen in in Table \ref{tab:noprune}, or data accumulation would be preferable, however our iterative approach naturally reinforces high-confidence, precise outputs.

\section{Conclusion}

This study presents an analytical investigation of  what factors influence self-improvement in language models, centering on our \textbf{Think, Prune, Train} framework. Our findings challenge the notion that simply scaling synthetic dataset size and model scale is sufficient to enhance reasoning capabilities.  We demonstrate that \textbf{correctness-based pruning} stabilizes training, preserving knowledge while enhancing reasoning. Our experiments highlight how recursive fine-tuning improves first-attempt accuracy, with \textbf{Gemma-2B increasing from 41.9\% to 57.6\% Pass@1, Gemma-9B reaching 82\%, and LLaMA-70B-Instruct achieving 91\%, surpassing GPT-4o’s 82\%}. Additionally, we identify key conditions for effective self-improvement, analyzing the impact of data volume, model size, pruning strategies, and fine-tuning iterations. This provides  insights into iterative reasoning refinement, showing that structured selection plays a more critical role than sheer dataset size. Finally, we evaluate the \textbf{Think, Prune, Train (TPT)} framework, demonstrating that structured reasoning prompting, correctness-based pruning, and supervised fine-tuning on validated solutions enable scalable self-improvement without external supervision, highlighting the potential of simplistic frameworks to unlock further advances in LLM reasoning and accuracy.

\section*{Acknowledgment}
We would like to thank Azalia Mirhoseini for her invaluable guidance and mentorship throughout this project. We are grateful to Anna Goldie for all her help and support and to Simon Guo for the helpful insights and guidance. We thank Rishabh Ranjan, Adrian Gamarra La Fuente, and the Scaling Intelligence Lab for creating such an inspiring environment. I also thank AMD for their GPU sponsorship that enabled the LLaMA 70B runs and Ceramic AI for access to their cluster for fine-tuning experiments.

\newpage

\bibliography{iclr2025_conference}
\bibliographystyle{iclr2025_conference}

\appendix
\newpage
\section{Appendix}

We have focused on GSM8K rather than more challenging datasets like GPQA because we consider smaller models with less memory and aim to study their reasoning capabilities rather than their recall ability. The \textit{reasoning} in more difficult problems is often not much more complicated than in grade school math but just covers more esoteric subjects. Consider a problem from GPQA \cite{rein2023gpqagraduatelevelgoogleproofqa} that asks about the angles of drops of liquid on a smooth and rough surface. We paraphrase the GPQA question here.

\begin{tcolorbox}[colback=yellow!10!white, colframe=pink!50!black, title=Complex Math Problem]
\begin{verbatim}
On a surface, we apply a coating which
makes it completely smooth. We put a 
drop of water and a hydrocarbon on it 
and measure the contact angles as 130° 
and 102°. We then roughen the surface
and measure the angle of water as 150°. 
What is the contact angle for a lighter 
hydrocarbon on the roughened surface? 
126°,131°,136°,141°
\end{verbatim}
\end{tcolorbox}

The Cassie-Baxter equation says that if $\theta_1$ is the contact angle on a smooth surface, then the angle on a rough surface, $\theta$, obeys $\cos(\theta) = f*\cos(\theta_1) + f$, where $f$ is a constant for the surface. The $f$ for the rough surface can be found from the two angles for water, $\cos(150°) = f*\cos(130°) - f$, so $f = 0.53$. Then, we can find the rough angle for the hydrocarbon, $\theta_h$, by using the formula $cos(\theta_h) = 0.53*\cos(105°) - 0.53$, so $\theta_h = 131°$. The other hydrocarbon is lighter, so it has lower surface tension; thus, its angle is less than $131$. The only choice that is less than $131°$ is $126°$.

This does not seem easy because few can readily recall (if ever knew) the formulas for the wettability of rough and flat surfaces. Once reminded of —the Cassie Baxter equation—the problem becomes isomorphic to the following.

Given two numbers, $x$, and $y$, and a formula $f$, work out another number, $z$, using the equation.
$x =f(y,z)$
Then, use $z$ and another instance of the formula to compute $w$ where
$w = f(y',z)$.
We are told $w'$ is less than $w$, and given only one choice that is less than $w$.
If $f$ is invertible then this is grade school math. 

The GPQA problem has the same type of reasoning as a grade school problem. The difference is using a obscure formulas that an LLM cannot recall.
\begin{tcolorbox}[colback=yellow!10!white, colframe=yellow!50!black, title=Simple Math Problem]
\begin{verbatim}
Jane has a number of buckets. 3 apples 
fit in each bucket.  She can fit 15 
apples in her buckets. If a bucket 
can fit 4 lemons, and limes are
smaller than lemons, how many limes 
can fit in her buckets?  
10, 15, 19, 24
\end{verbatim}
\end{tcolorbox}

GPQA questions often do not have complex reasoning but  are difficult due to the subject matter. As we aim to study reasoning, as opposed to the ability of the models to remember difficult facts, we believe GSM8K sufficient.

\section{Scaling Synthetic Data for Code Contest and LeetCode}

Our investigation into the impact of scaling synthetic code data reveals that increasing the amount of synthetic data does not always yield better performance. 

For example, in the Code Contest dataset, a Gemma-2B model trained with 2,000 self-generated and filtered synthetic examples outperformed the same model trained on synthetic data from Gemma-9B. However, as our ablation studies indicate, when unpruned, synthetic data from the 9B model leads to better performance than 2B-generated data.

This suggests that smaller models can generate high-quality data that, when validated effectively, can match or even surpass the usefulness of data from larger models. This trend is seen across datasets, reinforcing the importance of careful data curation and filtering.

\subsection{Diminishing Returns in Synthetic Data Scaling}

Pass@1 performance on Code Contest reveals an interesting trend: while synthetic data generated by Gemma-9B improves performance initially, further scaling results in stagnation or decline. At 4,000 and 6,000 synthetic examples, performance either levels off or worsens.

Similarly, introducing small amounts of human-generated training data negatively impacts Pass@1, possibly due to an out-of-distribution effect.

The LeetCode dataset shows a smaller performance gap between synthetic data generated by Gemma-2B and Gemma-9B. A model trained on 1,000 2B-generated examples achieved a Pass@1 of 14.65\%, slightly outperforming a model trained on 9B-generated examples, which achieved 12.82\%. Despite this, both models performed similarly on Pass@20 and Correct@100, again suggesting that smaller models can generate useful training data.

\subsection{Implications for Synthetic Data Generation Strategies}

These results demonstrate that simply increasing synthetic data volume does not guarantee better outcomes. Instead, there is a complex interplay between dataset size, data quality, and the model used for data generation.

Our findings highlight the need for targeted synthetic data generation strategies and reinforce the potential for smaller models to contribute meaningfully to model improvement.

\begin{table}[h]
\centering
\scriptsize
\setlength{\tabcolsep}{2pt}
\renewcommand{\arraystretch}{0.85}
\begin{tabular}{llcccc}
\toprule
\textbf{Model} & \textbf{Stage} & \textbf{Source} & \textbf{Data} & \textbf{Pass@1(\%)} & \textbf{Pass@50(\%)} \\
\midrule
\multirow{1}{*}{\textbf{Gemma-2B (Code Contest)}}  
& Baseline & - & - & 0.90 & 5.70  \\
& 1k-2B-synth & Gemma-2B & 1k & 0.84 & 7.14  \\
& 2k-2B-synth & Gemma-2B & 2k & 0.71 & 7.14  \\
& 4k-2B-synth & Gemma-2B & 4k & 0.56 & 7.85  \\
& 6k-2B-synth & Gemma-2B & 6k & 0.66 & 8.57  \\
\midrule
\multirow{1}{*}{\textbf{Human Data (Code Contest)}}  
& 1k-Human & Trainset & 1k & 0.60 & 6.43  \\
& 2k-Human & Trainset & 2k & 0.41 & 7.10  \\
& 4k-Human & Trainset & 4k & 0.24 & 5.71  \\
& 6k-Human & Trainset & 6k & 0.29 & 5.00  \\
\midrule
\multirow{1}{*}{\textbf{Gemma-9B (Code Contest)}}  
& Baseline & - & - & 5.20 & 15.00  \\
& 1k-9B-synth & Gemma-9B & 1k & 7.30 & 17.80  \\
& 2k-9B-synth & Gemma-9B & 2k & 7.10 & 18.50  \\
& 4k-9B-synth & Gemma-9B & 4k & 6.70 & 17.10  \\
\bottomrule
\end{tabular}
\caption{\textbf{Code Contest Scaling Amount of Synthetic Data Performance.} Pass@1 and Pass@50 performance on different sizes of synthetic data.}
\label{tab:code_contest_scaling}
\end{table}

\begin{table}[h]
\centering
\scriptsize
\setlength{\tabcolsep}{2pt}
\renewcommand{\arraystretch}{0.85}
\begin{tabular}{llcccc}
\toprule
\textbf{Model} & \textbf{Stage} & \textbf{Source} & \textbf{Data} & \textbf{Pass@1(\%)} & \textbf{Pass@20(\%)} \\
\midrule
\multirow{1}{*}{\textbf{Gemma-2B (LeetCode)}}  
& Baseline & - & - & 13.66 & 36.60  \\
& 1k-2B-synth & Gemma-2B & 1k & 14.65 & 33.66  \\
& 1k-9B-synth & Gemma-9B & 1k & 12.82 & 33.66  \\
\midrule
\multirow{1}{*}{\textbf{Gemma-9B (LeetCode)}}  
& Baseline & - & - & 13.66 & 36.60  \\
& 1k-9B-synth & Gemma-9B & 1k & 14.65 & 33.66  \\
\bottomrule
\end{tabular}
\caption{\textbf{LeetCode  Scaling Amount of Synthetic Data Performance.} Pass@1 and Pass@20 performance with different sizes of synthetic data.}
\label{tab:leetcode_scaling}
\end{table}

\subsection{Impact of Mixing Real and Synthetic Data}

We also explored the impact of combining small amounts of real and synthetic data for training. Our findings indicate that mixed datasets performed well in terms of Pass@1 scores, particularly in smaller training sizes (e.g., \texttt{1k 500 each}). However, attempts to recursively refine models using mixed datasets consistently failed. A possible explanation is that the introduction of real data disrupts the recursive self-improvement process, making it more difficult for the model to generalize effectively from its own outputs.

\begin{table}[h]
\centering
\scriptsize 
\setlength{\tabcolsep}{2pt} 
\renewcommand{\arraystretch}{0.85} 
\begin{tabular}{llcccc}
\toprule
\textbf{Model} & \textbf{Stage} & \textbf{Source} & \textbf{Data} & \textbf{Pass@1(\%)} & \textbf{Pass@50(\%)} \\
\midrule
\multirow{1}{*}{\textbf{Gemma-2B}} \\ (Code Contest, Mixed)  
& 1k-500-each & Trainset/2B & 1k & 1.0 & 7.14  \\
& 2k-1k-each & Trainset/2B & 2k & 0.60 & 7.85  \\
& 4k-2k-each & Trainset/2B & 4k & 0.24 & 5.71  \\
\midrule
\multirow{1}{*}{\textbf{rec mixed-sft}}  
& 1k both on 1k both & 1kboth & 1k & 0.87 & 7.14  \\
& 2B1k on base & Gemma-2B & 1k & 0.92 & 4.28  \\
\bottomrule
\end{tabular}
\caption{\textbf{Mixed Dataset Ablations.} Evaluating model performance when fine-tuning on a mix of real and synthetic data.}
\label{tab:mixed_dataset_ablation}
\end{table}

\subsection{The CoT prompt used}

The following is the Chain-of-Thought (CoT) prompt used in our experiments.

\label{appendix-prompt}
\begin{tcolorbox}[colback=blue!5!white, colframe=blue!75!black, title=\tiny{Mathematical Reasoning Prompt}]
\tiny
\textbf{You are an expert mathematician.}  

You are provided with a math problem.  
Your task is to solve the problem step-by-step, clearly showing all relevant calculations and reasoning.  

\vspace{0.1cm}
\noindent\textbf{Problem:}  
\texttt{"Bella bought stamps at the post office. Some of the stamps had a snowflake design, some had a truck design, and some had a rose design. Bella bought 11 snowflake stamps. She bought 9 more truck stamps than snowflake stamps, and 13 fewer rose stamps than truck stamps. How many stamps did Bella buy in all?"}

\vspace{0.1cm}
\noindent\textbf{Requirements:}
\begin{enumerate}
    \item Provide a complete and correct solution in a markdown block.
    \item Explain each step of the solution in detail.
    \item Conclude with the final numerical answer on a new line in the format \texttt{\#\#\#\# [Answer]}, replacing \texttt{[Answer]} with the actual answer.
\end{enumerate}

\vspace{0.1cm}
\noindent\textbf{Solution:}  
\end{tcolorbox}

\subsection{Example reasoning traces}

\label{appendix-traces}
\begin{tcolorbox}[colback=blue!5!white, colframe=blue!75!black, title=Gemma 2B]
\tiny
\textbf{Solution:}
\texttt{Here's a step-by-step solution to the problem: \\
        **1. Calculate the number of truck stamps:** \\
        Bella bought 9 more truck stamps than snowflake stamps, and she bought 11 snowflake stamps.\\
        Therefore, she bought 11 + 9 = 20 truck stamps. \\
        **2. Calculate the number of rose stamps:** \\
        Bella bought 13 fewer rose stamps than truck stamps, and she bought 20 truck stamps.\\
        Therefore, she bought 20 - 13 = 7 rose stamps.\\
        **3. Calculate the total number of stamps:** \\
        Bella bought 11 snowflake stamps + 20 truck stamps + 7 rose stamps = 38 stamps in total.\\
        **\#\#\#\# 38**}
\end{tcolorbox}

\begin{tcolorbox}[colback=green!5!white, colframe=green!75!black, title=Gemma 9B]
\tiny
\textbf{Solution:}
\texttt{
Here's the step-by-step solution to the problem:\\
1.**Find the number of truck stamps:** \\
* Bella bought 9 more truck stamps than snowflake stamps, which she bought 11 of.\\
* Therefore, she bought 11 + 9 = 20 truck stamps.\\
2. **Find the number of rose stamps:**\\   
* Bella bought 13 fewer rose stamps than truck stamps, which she bought 20 of.\\
* Therefore, she bought 20 - 13 = 7 rose stamps.\\
3. **Calculate the total number of stamps:** \\
* To find the total number of stamps, add the number of each type of stamp: 11 snowflake stamps + 20 truck stamps + 7 rose stamps = 38 stamps.\\ 
\#\#\#\# 38}
\end{tcolorbox}

\begin{tcolorbox}[colback=red!5!white, colframe=red!75!black, title=LLaMA 70B]
\tiny
\textbf{Solution:}
\texttt{
\#\#\# Step-by-step solution to the problem\\ 
To find the total number of stamps Bella bought, we first need to determine the number of stamps for each design. We are given the following information:\\
- The number of snowflake stamps = 11\\
- The number of truck stamps = 9 more than snowflake stamps\\
- The number of rose stamps = 13 fewer than truck stamps\\
\#\#\#  Calculating the number of truck stamps\\
Since Bella bought 9 more truck stamps than snowflake stamps, we calculate the number of truck stamps by adding 9 to the number of snowflake stamps.\\
Number of truck stamps = Number of snowflake stamps + 9\\
= 11 + 9\\
= 20\\
\#\#\#  Calculating the number of rose stamps\\
Given that Bella bought 13 fewer rose stamps than truck stamps, we calculate the number of rose stamps by subtracting 13 from the number of truck stamps.\\
Number of rose stamps = Number of truck stamps - 13\\
= 20 - 13\\
= 7\\
\#\#\#  Calculating the total number of stamps\\
To find the total number of stamps, we add the number of snowflake, truck, and rose stamps together.\\
Total number of stamps = Number of snowflake stamps + Number of truck stamps + Number of rose stamps\\
= 11 + 20 + 7\\
= 38\\
\#\#\#\# 38}
\end{tcolorbox}
\hfill

\begin{tcolorbox}[colback=yellow!5!white, colframe=yellow!75!black, title=LLaMA 1B]
\tiny
\textbf{Solution:}
\texttt{
\#\#\# Step 1: Interpret the Problem and Identify Given Information\\
\#\#\# Step 2: Set Up Equations Based on Given Information\\
From the problem, we can set up the following equations:\\
1. `s = 11` (number of snowflake stamps)\\
2. `t = s + 9` (number of truck stamps)\\
3. `r = t - 13` (number of rose stamps)\\
\#\#\# Step 3: Express the Number of Rose Stamps in Terms of Snowflake Stamps and Truck Stamps\\
Now, we'll substitute the expressions for `t` and `r` into the third equation:\\
`s = 11`\\
`t = s + 9`\\
`r = (s + 9) - 13`\\
\#\#\# Step 4: Solve for the Number of Rose Stamps\\
Let's simplify the expression for `r`:\\
`r = (s + 9) - 13`\\
`r = s - 4`\\
So, the number of rose stamps in terms of the number of snowflake stamps is `r = s - 4`.\\
\#\#\# Step 5: Find the Total Number of Stamps\\
The total number of stamps is the sum of snowflake stamps, truck stamps, and rose stamps. Therefore, we can write an equation representing the total number of stamps:\\
`s + t + r = 11 + (11 + 9) + (s - 4)`\\
Simplify and solve for `s` (snowflake stamps):\\
`s + t + r = 11 + 20 + s - 4`\\
`s + t + r = 26 + s`\\
`s - s = 26`\\
`s = 26`\\
Since `s` represents the number of snowflake stamps and we know `s = 11`, the actual number of snowflake stamps is indeed 11. Let's confirm this with the other equations.\\
\#\#\# Step 6: Confirm the Number of Truck and Rose Stamps\\
Now, we'll find the number of truck stamps `t` and rose stamps `r` using the equations:\\
`t = s + 9`\\
`r = t - 13`\\
We found `s = 11`, so:\\
`t = 11 + 9`\\
`t = 20`\\
`r = 20 - 13`\\
`r = 7`\\
\#\# Step 7: Verify the Total Number of Stamps Calculations\\
To ensure the correctness of our solution:\\
`s + t + r = 11 + 20 + 7`\\
`s + t + r = 38`\\
Using the original formula `20 + 7 - 4 = 37`, this confirms our conclusion that `s = 11` is a perfect and easy-to-explain solution that satisfies all given conditions.\\
\#\#\# Step 8: Conclusion with the Final Answer\\
We finally have the total number of stamps Bella bought in all - 11 snowflake stamps, 20 truck stamps, and 7 rose stamps.\\
\#\#\#\# 38
}
\end{tcolorbox}

\end{document}